\documentclass[letterpaper, 10 pt, conference]{ieeeconf}  % Comment this line out
\IEEEoverridecommandlockouts                              % This command is only
\usepackage{mathptmx} % assumes new font selection scheme installed
\usepackage{amsmath} % assumes amsmath package installed
\usepackage{amssymb}  % assumes amsmath package installed
\usepackage[linesnumbered,ruled,vlined]{algorithm2e}
\usepackage{tabularx}
\usepackage{graphicx} 
\usepackage{booktabs}
\usepackage{multirow}
\usepackage{multicol}
\usepackage{xcolor}
\usepackage{MnSymbol}
\usepackage[hidelinks]{hyperref}
\title{\LARGE \bf
An Active Search Strategy with Multiple Unmanned Aerial Systems for Multiple Targets
}

\author{Chuanxiang Gao, Xinyi Wang, Xi Chen, and Ben M. Chen
\thanks{This project is supported in part by the Research Grants Council of Hong Kong SAR (Grant No: 14209020 and Grant No: 14206821) and in part by the InnoHK of the Government of Hong Kong via the Hong
Kong Centre for Logistics Robotics.}% <-this % stops a space
\thanks{The authors are with the Department of Mechanical and
Automation Engineering, The Chinese University of Hong Kong, Shatin,
N.T., Hong Kong. E-mail: 
        {\tt\small cxgao@mae.cuhk.edu.hk, xywangmae@link.cuhk.edu.hk, \{xichen002,bmchen\}@cuhk.edu.hk}}%
}

\begin{document}

\maketitle
\thispagestyle{empty}
\pagestyle{empty}

%%%%%%%%%%%%%%%%%%%%%%%%%%%%%%%%%%%%%%%%%%%%%%%%%%%%%%%%%%%%%%%%%%%%%%%%%%%%%%%%
\begin{abstract}

The challenge of efficient target searching in vast natural environments has driven the need for advanced multi-UAV active search strategies. This paper introduces a novel method in which global and local information is adeptly merged to avoid issues such as myopia and redundant back-and-forth movements. In addition, a trajectory generation method is used to ensure the search pattern within continuous space. To further optimize multi-agent cooperation, the Voronoi partition technique is employed, ensuring a reduction in repetitive flight patterns and making the control of multiple agents in a decentralized way. Through a series of experiments, the evaluation and comparison results demonstrate the efficiency of our approach in various environments. The primary application of this innovative approach is demonstrated in the search for horseshoe crabs within their wild habitats, showcasing its potential to revolutionize ecological survey and conservation efforts.
\end{abstract}

%%%%%%%%%%%%%%%%%%%%%%%%%%%%%%%%%%%%%%%%%%%%%%%%%%%%%%%%%%%%%%%%%%%%%%%%%%%%%%%%
\section{Introduction}

Unlike traditional search methodologies that passively analyze data, active search is a dynamic process where the searcher actively selects which data points to query next based on previous observations. The goal is to maximize the efficiency of the search by focusing on the most informative or valuable data points. It has many robotics applications, such as environment monitoring and search and rescue \cite{Bouras2022, Martinez2022}. The realm of active search problems, especially in expansive and unpredictable environments, has long been a topic of interest and research \cite{Flaspohler2019, Nawaz2023, Igoe2022}. It can be regarded as the extension of coverage path planning, information gathering, and exploration.

Coverage path planning (CPP) is concerned with determining a path for a searcher (or multiple searchers) such that every point in a designated space is covered or visited. In \cite{Manjanna2017}, the authors present a methodology for marine vehicles to cover a region of the sea surface by using a selective sampling method. By leveraging adaptive sampling, their technique can exploit environmental clustering phenomena, resulting in a more accurate representation of the marine field where traditional sampling methods like boustrophedon or lawnmower path may fall short. For the multi-agent coverage path planning problem, Athanasios et al. \cite{Kapoutsis2017} propose a novel algorithm that aims to converge to the optimal solution. This is achieved by transforming the original integer programming problem of the multi-agent CPP problem into several single-robot CPP problems, which effectively bypass the combinatorial complexity of the original multi-agent CPP problem. However, standard CPP algorithms do not adapt to new information, like potential locations of targets or areas of high probability, which would waste time and resources covering areas where targets are unlikely to be found.

Although CPP ensures comprehensive coverage, information gathering focuses on the quality and relevance of collected data \cite{Schwager2017, Khan2015}. 
Efficient information gathering can significantly improve active search outcomes. By collecting relevant and high-quality data, the agent can make better-informed decisions, increasing the chances of successfully locating its targets. In \cite{Hollinger2014}, the authors utilize sampling-based methods with information gain and present a novel framework that enables robots to efficiently gather information from their surroundings. Based on this idea, Michael et al. \cite{Kuhlman2017} aim to optimize the search process by considering both the current state of the environment and potential future changes. This paper uses mutual information as a metric to guide search strategies, increasing the likelihood of successfully locating targets.

Furthermore, exploration and active search are closely related. While exploration focuses on investigating or navigating an environment without a specific target in mind \cite{Cao2021, Dharmadhikari2020, Dharmadhikari2023}, active search involves a purposeful and directed effort to find specific targets or features in an environment. Exploration is essential when the environment is unknown or partially known. Frontier-based exploration \cite{Cieslewski2017,Lu2020} methods are widely used to guarantee the global coverage of unknown environments. On the contrary, the next-best-view strategy \cite{Bircher2016} focuses on exploring the surrounding regions and has good performance in reducing the path length. The approach described in this paper is based on key results from these three research areas.

In this paper, we focus on the multi-agent continuous active search for unknown multiple targets in the environment. The proposed approach can handle uncertainties in the environment. Only a coarse prior map is needed to guide the search process. Both local and global information are used, and the movement process is formulated into a Markov Decision Process (MDP) to determine the next and future actions. The strength of our method lies in its versatility and efficiency. Not only does it ensure that every point is covered, but it also guarantees that the information gathered is both deep and relevant. The inclusion of the Voronoi partition technique further reduces the repetitive flight area in multi-agent scenarios. The proposed approach can achieve fast and safe search and find multiple targets without knowing their locations. The major contributions of this paper are summarized as follows:

\begin{itemize}
    \item  We propose a continuous active search strategy for multiple targets based on coarse probabilistic information, which considers global and local information to avoid myopia problems and redundant movements.
    \item  We extend the active search with a single UAV to a cooperative search with multiple UAVs. Voronoi partition is employed to reduce the overlap and handle the safety of each UAV in a decentralized way.
    \item  We apply the proposed approach in a real application scenario, which involves searching and finding horseshoe crabs within their habitats.
\end{itemize}

The remainder of this paper is organized as follows. In Section~\ref{sec: Related}, we give some related works about active search strategy for multi-target. Section~\ref{sec: Formulation} shows the problem formulation of cooperative search with multiple UAVs. The proposed single-UAV and multi-UAV search strategy is presented in Section~\ref{sec: Method}. Section~\ref{sec: Evaluations} presents the evaluations and comparisons. Conclusions are presented in Section~\ref{sec: Conclusion}.

% \textbf{Notation.} $\mathbb{R}$ denotes the set of real numbers. ${\rm T}$ denotes transpose of matrix. For a set $a$, ${\rm card}(a)$ denotes the number of elements of $a$.

%%%%%%%%%%%%%%%%%%%%%%%%%%%%%%%%%%%%%%%%%%%%%%%%%%%%%%%%%%%%%%%%%%%%%%%%%%%%%%%%
\section{Related Works}
\label{sec: Related}
The search strategy originates from the foraging behavior of animals. One of the most typical search patterns is spiral \cite{Gelenbe1997}, designed to reduce the time needed to locate a single stationary target. For the active search with a single UAV, three pivotal search strategies are analyzed: Global maxima search, Local maxima search, and Spiral search \cite{Meghjani2016}. The authors aim to strike a balance between minimizing the mean time to find a target and maximizing the likelihood of its discovery.  To maximize the information value of data acquired in an initially unknown environment, Rückin et al. \cite{Rückin2022} integrate tree search with an offline-trained neural network. This neural network serves to predict informative sensing actions. In addition, Saulnier et al. \cite{Saulnier2020} combine a deterministic tree search algorithm and an efficient long-horizon uncertainty prediction method to optimize the potential sensing trajectories.

For multi-agent active search problems, Manjanna et al. \cite{Manjanna2022} divide them into several single-agent active search problems and use MDP to model the movement of each agent. The reward is designed according to the probabilistic map and the distance between the target and the agent. Instead of decoupling the multi-agent search problem, Yao et al. \cite{Yao2021} propose a bidirectional negotiation method to determine the subregions that each agent needs to search. Each agent follows the adaptive spiral coverage strategy to cover each sub-region. Taking into account communication constraints, the authors \cite{Ebert2022} introduce a two-stage hybrid algorithm: the first stage involves the location of a target using a variation of PSO, while the second stage focuses on maximizing the dissemination of the target knowledge across the robot group. With regard to detection uncertainty, a noise-aware Thompson sampling algorithm, designed for multiple ground-based robots, is introduced to address this problem \cite{Ghods2021}.
However, the aforementioned search strategies rely on either local or global information exclusively. In contrast, our focus lies in the development of a highly efficient multi-agent continuous search strategy that aims to overcome myopia and prevent repetitive flight.
\section{Problem Definition}
\label{sec: Formulation}

\begin{figure}[!htb]
  \centering
  \includegraphics[width=0.8\hsize]{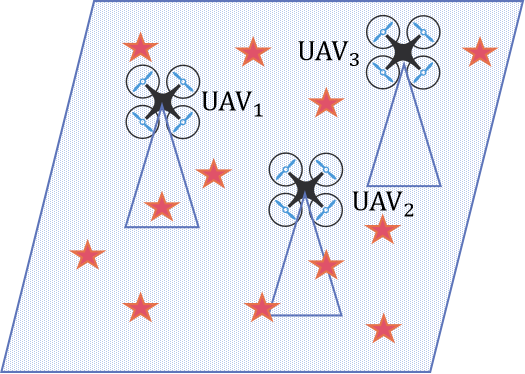}
  \caption{Illustration of the search problem. The red stars represent the target objects that are randomly located at the map.}
  \label{fig: environment}
\end{figure}

The overall problem described in this paper is using multiple UAVs to cooperatively search unknown locations of targets in a wild environment as shown in Fig. \ref{fig: environment}. Define $\mathbf{M} \in \mathbb{R}^2$ as the workspace to be searched. Let $\rm N$ be the total number of target objects $\mathcal{O}$. The precise localization of the targets is unknown. The probabilistic distribution $\mathbf{\mathcal{P}}$ that reflects the possible occurrence rate is given. Consider a team of $n$ UAVs located at $\mathbf{P}(t) = [\mathbf{P}_1(t),\dots,\mathbf{P}_n(t)], \mathbf{P}(t) \in \mathbf{M}$. Each UAV is equipped with a downward-facing camera with 90$^\circ$ field of view (FOV) and flies at a certain distance above the ground. As a result, each UAV can be regarded as a UAV moving in the workplace $\mathbf{M}$ with a range of sensors $\mathbf{S}$. The sensor range of each UAV is assumed to be the same. The trajectories of UAVs $\mathbf{J}(t) = [\mathbf{J}_1(t),\dots,\mathbf{J}_n(t)]$ follows kinodynamic constraints. We assume the detection rate is 1, which means that if the object $\mathcal{O}_i, i=1,\dots,\rm N$ is in the sensor area where the UAVs are located, it is assumed that the object is detected. The problems can be defined as follows.
\begin{itemize}
    \item \emph{Problem 1}: Within a certain search time $T$, maximize the total number of objects to be found, where each object $\mathcal{O}_i$ can only be found once.
    \item \emph{Problem 2}: Cover all the targets, minimize the maximum value of the distance of the trajectories $\mathbf{J}$, where the trajectories must be kinodynamically feasible.
\end{itemize}

\section{Methodology}
\label{sec: Method}
The proposed method for cooperative search with multiple UAVs utilizes a combination of local information, global information, and the correlation relationship among UAVs to achieve a rapid and efficient continuous active search for multiple targets. This section provides a detailed explanation of the proposed method.
\subsection{Active Search with a Single UAV}
\label{sec: MethodSingle}
In the context of active search with a single UAV, our method employs a search tree that effectively captures the various possibilities for local movements.	
The goal is to design a movement strategy that maximizes information gain. 
By incorporating global information, the growth of the search tree is intricately linked to the understanding of the environment. As a result, the search tree preferentially expands towards areas rich in valuable information.
This integration between local movements and global information enables our method to significantly enhance search efficiency and maximize the exploration of informative areas within the environment.

\begin{figure}[!htb]
  \centering
  \includegraphics[width=\hsize]{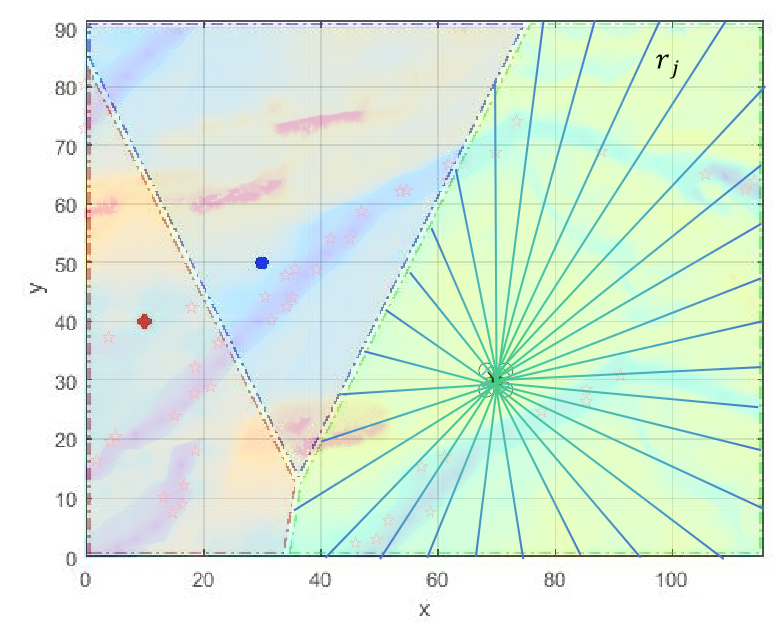}
  \caption{Illustration of sampling rays. Voronoi partition is used to bind the rays.}
  \label{figure: global}
\end{figure}

Define $S$ as all the discrete directions shown in Fig. \ref{figure: global}.
One circumference is uniformly discrete into $N_s$ rays $r$ with the localization of the UAV $\mathbf{P}_i(t)$ as the starting point. The information gain for each ray $r_j, j=1,\cdots,N_s$ can be defined as follows:
\begin{equation}
    I(S_j) = \frac{\mathbf{\mathcal{P}}(r_j)}{\sum_{k=1}^{N_s} \mathbf{\mathcal{P}}(r_k)}
\end{equation}
where $\mathbf{\mathcal{P}}(r_j)$ is the sum value of the grids that each ray $r_j$ passes. This normalization of information gain facilitates the identification of the direction with the maximum information yield. The discretization into uniform rays dictates the local search tree's expansion direction. During the expansion process, the direction with higher information gain will likely be chosen. The uniformly discrete process is designed to make full use of global information to guide the movement of UAVs.

The local movement of the UAV is characterized as uniformly accelerated motion.
Let $\mathbf{V}_i(t)$ represent the velocity of the $i$th UAV and 
$\mathbf{V}_s$ denote the termination velocity.
\begin{equation}
    \label{eq: velocity}
    \mathbf{V}_s = [V_{\max}\cos(S_i), V_{\max}\sin(S_i)]
\end{equation}
$V_{\max}$ is the maximum velocity of UAVs. Over a time interval $T$, the position of the UAV can be updated as follows:
\begin{equation}
    \label{eq: simulation}
    \mathbf{P}_i(t+T) = \mathbf{P}_i(t) + \frac{\mathbf{V}_i(t)+\mathbf{V}_s}{2}
\end{equation}
Furthermore, the expansion of the search tree is formulated as a MDP. By anticipating $N$ steps forward, we aim to identify the optimal policy $\pi^\star$ that maximizes the expected cumulative reward: 
\begin{equation}
    \label{eq: dp}
    \pi^\star = \arg\max_\pi \mathbb{E} \left[ \sum_{t}^{NT} \gamma^t R(\mathbf{P}_i(t), a_t) \right]
\end{equation}
To enhance global and local information utilization, the action space $a_t \in S$ dynamically changes at each step, following a specifically designed process. Initially, at each step, $N_a$ number of directions are selected from $S$ based on their information gain. Subsequently, for each direction, $S_i, i=1,\cdots, N_a$, a forward simulation is performed based on Eq.~\eqref{eq: simulation}, and the trajectory $a^i_t$ from time $t$ to time $t+T$ constitutes one action. Consequently, the action space $a_t$ comprises all possible trajectories $a^i_t$. The reward $R(\mathbf{P}_i(t), a_t)$ is determined by the cumulative value of the grids traversed by each trajectory. The state transition rate is assumed to be one where the transition is based on Eq.~\eqref{eq: simulation}. Finally, the optimal policy $\pi^\star$ that represents a sequence of actions with the highest information gain can be calculated by solving a dynamic programming problem.
% \begin{figure}[!htb]
%   \centering
%   \includegraphics[width=\hsize]{global.png}
%   \caption{Illustration of expansion of local search tree.}
%   \label{figure: local}
% \end{figure}
Upon determining the optimal policy, the terminal position $\mathbf{P}_i(t+NT)$ is defined. Building on our prior research \cite{Gao2023}, the transition from the current to the terminal position is modeled as an energy-efficient trajectory optimization process, represented by a fifth-order polynomial function as outlined in Eq.~\eqref{eq: poly}:
\begin{equation}
    \label{eq: poly}
    \mathbf{J}_i(t)= \sum_{k=0}^{5}\lambda_{k}t^k
\end{equation}
The optimization problem is formulated as follows:
\begin{equation}
\label{eq: traj}
\begin{aligned}
\min~~&E(\mathbf{J}_i(t))\\
\mbox{s.t.}~~&\mathbf{J}_i(t+NT) = \mathbf{P}_i(t+NT) \quad \\
&\mathbf{J}_i(t) = \mathbf{P}_i(t) \quad\\
&\mathbf{J}_i'(t) = \mathbf{V}_i(t) \quad\\
&\mathbf{J}_i''(t) = \mathbf{A}_i(t) \quad\\
&0 < \|\mathbf{J}_i'(t)\| < V_{\max}, \quad \quad \forall t \in [t,t+NT]\\
&0 < \|\mathbf{J}_i''(t)\| < A_{\max}, \quad \quad \forall t \in [t,t+NT]\\
\end{aligned}
\end{equation}
where $E(\mathbf{J}_i(t))$ is the energy cost of the trajectory, $\mathbf{A}_i(t)$ is the current acceleration of the UAV, $A_{\max}$ is the maximum acceleration allowed. The first constraint guarantees the UAV stops at the termination position, while constraints 2-4 ensure that the UAV's movement is continuous. The last two constraints ensure the velocity and acceleration are feasible. The above optimization problem can be rewritten into quadratic programming and solved efficiently.
\subsection{Cooperative Search with Multiple UAVs}
\label{sec: Multi}
Taking into account the instability of communication, we implement a decentralized cooperative method with multiple UAVs. Each UAV receives only the localization information from other UAVs and makes the decision itself. To improve the efficiency of searching and avoid repetitive searches for the same area, we use the Voronoi partition to divide the whole environment uniformly into non-intersecting regions for each UAV. The Voronoi partition is defined as follows:
\begin{equation}
\begin{aligned}
\label{voronoi}
&\mathcal{V}_i = \{\mathbf{p} \in \mathbf{M} | \Vert \mathbf{p}-\mathbf{P}_i(t)\Vert \leq  \Vert \mathbf{p}-\mathbf{P}_j(t)\Vert\ , \forall j\neq i\}.
\end{aligned} 
\end{equation}
This equation is used to guarantee that the distance between any point and the UAV is closer than the distance between this point and other UAVs, which means each UAV $i$ is responsible for searching within its corresponding Voronoi region $\mathcal{V}_i$. The partitioning process is dynamic and real-time according to the positions of UAVs to prevent multiple UAVs from searching for the same region.

Algorithm \ref{algo: multi} presents the overall process of the multi-UAV cooperative active search method. Given the workspace $\mathbf{M}$ and objects $\mathcal{O}$, $n$ UAVs are used to search for these objects. The number of founded objects $N_f$ is first set to 0. At each iteration, the Voronoi partition is performed according to the position $\mathbf{P}$ of UAVs (lines 3-4). The whole workspace is divided into several subregions $\mathcal{V}_i$, where each region represents the best workspace of the $i$th UAV. For each UAV, the probabilistic map $\mathcal{P}$ with the highest value in its region is extracted as the global maxima position $\mathbf{P}_G$ (lines 5-6). A look ahead process is formulated into a dynamic programming problem based on Eq.~\eqref{eq: velocity}-\eqref{eq: dp} to get the best local position $\mathbf{P}_L$ (line 7). The UAV treats the best local position as the goal $\mathbf{G}$ if it is founded; otherwise, it treats the best global position as the goal (lines 8-11). To make the movement continuous, trajectory optimization based on Eq.~\eqref{eq: poly}-\eqref{eq: traj} generates the energy-optimal trajectory. During UAV movement, the number of objects found is updated according to the sensor ranges of the UAV $\mathbf{S}$ (line 13). The probabilistic map $\mathcal{P}$ is updated at each iteration based on the position of UAVs (line 14).
\begin{algorithm}[htbp]
    \caption{Multi-UAV Active Search Algorithm}
    \label{algo: multi}
    \KwIn{$\mathbf{M}, n, \mathcal{O}, \mathcal{P}, \mathbf{S}, \rm N, step$}
    $N_f \leftarrow 0$\;
    \While{$N_f\leq$ \rm N}{
        $\mathbf{P}\leftarrow GetPosition(n)$\;
        $\mathcal{V}\leftarrow GetVoronoiPartition(\mathbf{P})$\;
        \ForEach{\rm region $\mathcal{V}_i \in \mathcal{V}$}{
            $\mathbf{P}_G\leftarrow GetGlobalMaxima(\mathcal{V}_i,\mathcal{P})$\;
            $\mathbf{P}_L\leftarrow Lookahead(\mathcal{V}_i,\mathbf{P}_i,\mathcal{P},\rm step)$\;
            \eIf{$\mathbf{P}_L=\{ \}$}{
            $\mathbf{G}\leftarrow \mathbf{P}_G$\; 
            }
            {
            $\mathbf{G}\leftarrow \mathbf{P}_L$\;}
            $\mathbf{J}_i\leftarrow TrajectoryGeneration(\mathbf{P}_i,\mathbf{G})$}
        $N_f\leftarrow UpdateFoundedObjects(\mathbf{P},\mathcal{O})$\;
        $\mathcal{P}\leftarrow UpdateMap(\mathcal{P},\mathbf{P})$\;
        }

\end{algorithm}

\section{Evaluation and comparison}
\label{sec: Evaluations}
\begin{figure}[!htb]
  \centering
  \includegraphics[width=\hsize]{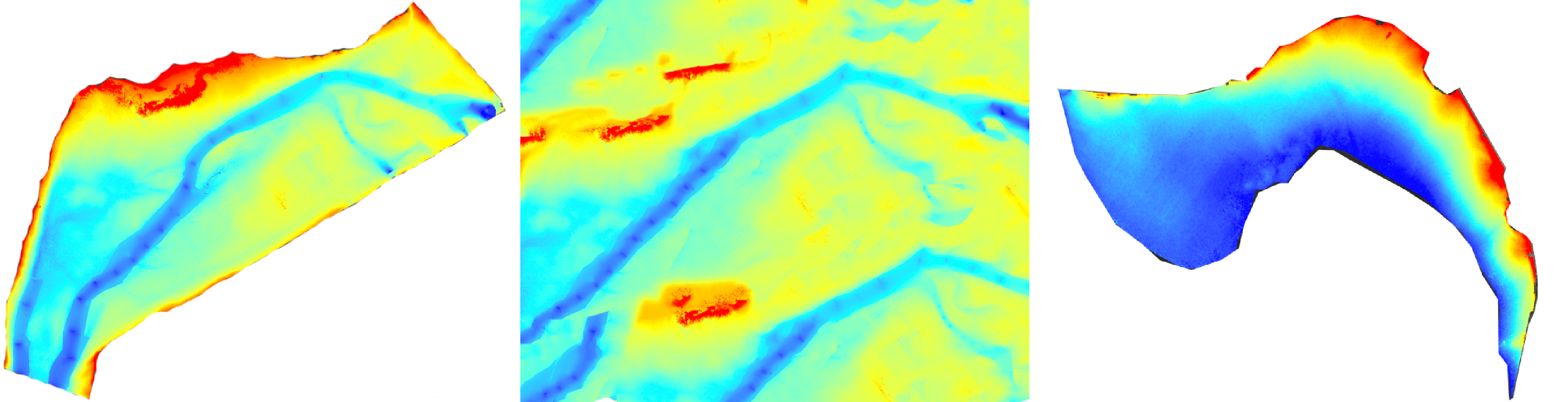}
  \caption{Illustration of simulation environments. The map with different colors represents the change in height.}
  \label{figure: experiment}
\end{figure}

In this section, we implement the proposed algorithm in simulated environments based on the data from \cite{Koyama2020}. The simulation environments are digital elevation maps, as shown in Fig. \ref{figure: experiment}. We take the horseshoe crab search as an example. The relationship between horseshoe crab occurrence and elevation changes follows a Poisson distribution, which serves as a suitable example for our analysis. As a result, a rough probabilistic map can be generated to guide the UAV, and the real distribution is used as the ground truth. It is noted that the horseshoe crabs search problem is just an example. The proposed algorithm can solve any search problems with a prior rough map. In these experiments, the lookahead step is set as 5, the maximum velocity $V_{\max}$ is set as 2$m/s$, the maximum acceleration $A_{\max}$ is set as 2$m/s^2$, the sensor range $\mathbf{S}$ is set as a square with a side length of 5 meters.

\subsection{Comparison of Search Strategies with a Single UAV}
We first evaluate the proposed active search strategy with a single UAV by comparing four search algorithms. Zig-zig search is a full coverage search strategy without any preference. Global Maxima (GM) search always chooses to visit the location with the highest information value in the map until the targets are found or the map is covered \cite{Meghjani2016}. Local Maxima (LM) search seeks to visit the location with the highest information value that is near the UAV \cite{Meghjani2016}. The heuristic local maxima (HLM) search shares the same search strategy as the local maxima search, but it can increase the search scope when the UAV is stuck. We compare the path length of different algorithms in the three scenarios shown in Fig. \ref{figure: experiment}. The UAV needs to find all the targets based on the prior maps. The comparison result is shown in Table \ref{table: single}. The path length of the Zig-zig search is the longest because it has no preference. The path length of the Global Maxima search is longer than the Heuristic Local Maxima search because it only considers the global information, which results in many back and forth. It can be seen that our method outperforms the other search strategies in the path length, as our method combines both local and global information. 

\begin{table}[htbp!]
\caption{\textbf{Quantitative comparison results on three scenarios}}
\label{table: single}
\begin{center}
\resizebox{0.8\columnwidth}{!}{\begin{tabular}{l c c c}
\toprule
\multirow{2}{*}{Methods} & \multicolumn{3}{c}{Path length (m)}\\\cline{2-4}
& Scene 1 & Scene 2  & Scene 3 \\
\midrule
Zig-zig & 3796 & 6828 & 1072\\
GM & 1845.9 & 2650.7 & 1212\\
LM & - & - & -\\
HLM & 1520.6 & 1171.4 & 921.4\\
\hline
Ours & \textbf{828.1} & \textbf{1060.2} & \textbf{536.3}\\

\bottomrule

\end{tabular}}
\end{center}
\vspace{-1mm}
\end{table}

\begin{figure}[!htb]
  \centering
  \includegraphics[width=\hsize]{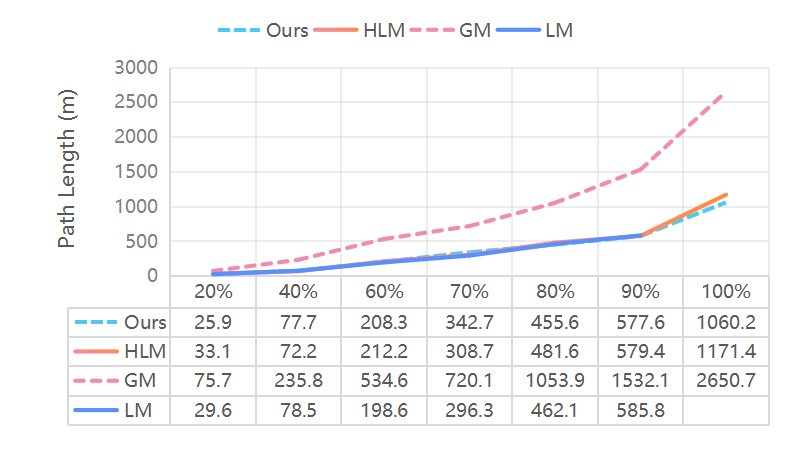}
  \caption{Comparison results of different algorithms when finding different numbers of targets.}
  \label{figure: successrate}
\end{figure}

As the Local Maxima search is always stuck in a local minimum, we compare the path lengths when different numbers of targets are found. As shown in Fig. \ref{figure: successrate}, the HLM and LM methods perform well when the number of discovered targets amounts to less than 70 percent of the total. This is due to their property of choosing nearby optimal steps. Nevertheless, the LM search method demonstrates a significantly low success rate in locating all the targets, resulting in the ineffective accomplishment of the complete search task. In contrast, our method showcases exceptional search capabilities throughout the entire process and remains unimpeded by local minima, thereby ensuring full coverage of the targets.

\subsection{Comparison of Search Strategies with Multiple UAVs}
\begin{figure}[!htb]
  \centering
  \includegraphics[width=\hsize]{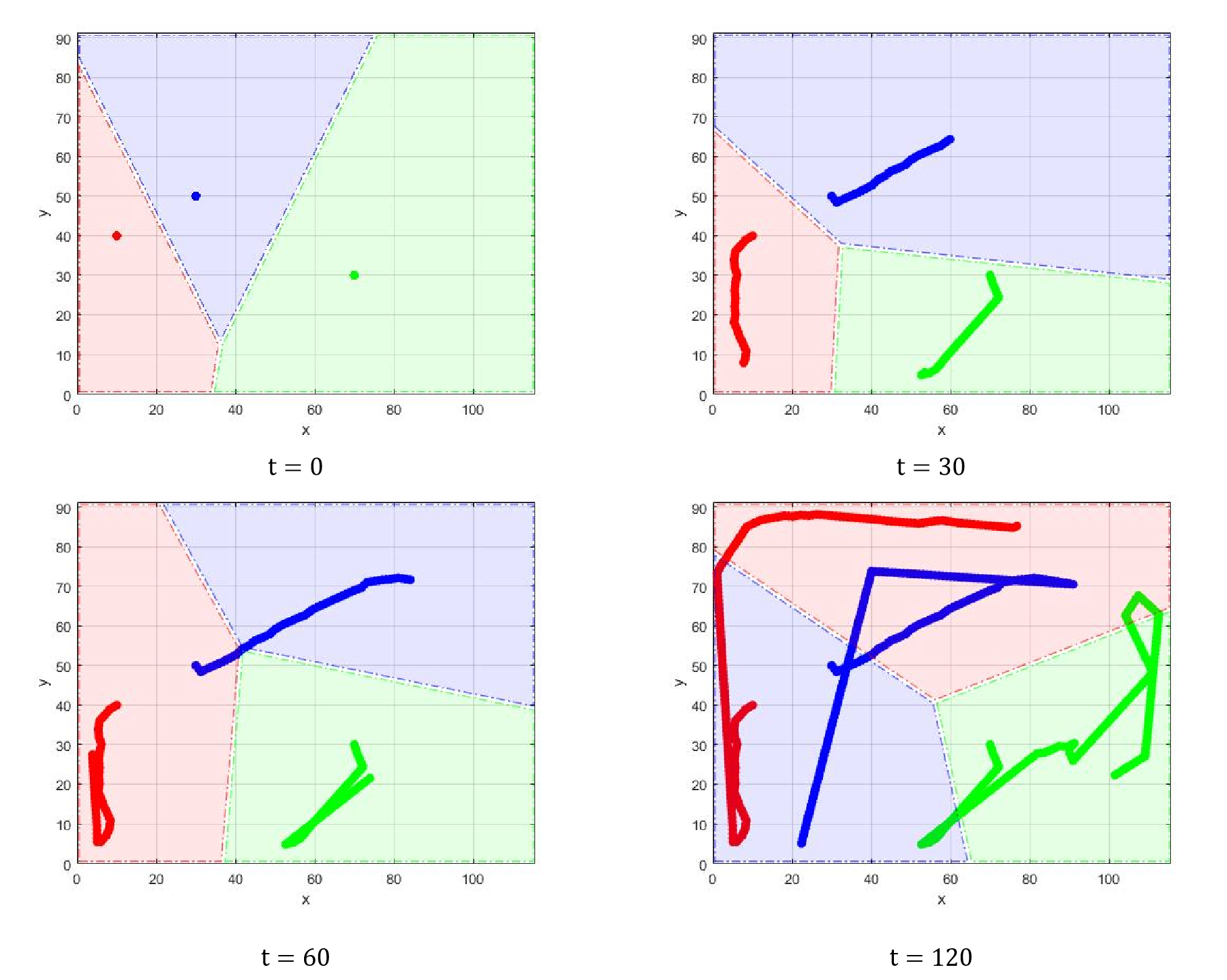}
  \caption{Illustration of the update of region division and the path of each UAV.}
  \label{figure: multi}
\end{figure}

To assess the effectiveness of our method in the aforementioned scenarios, we conduct performance evaluations. The objective of the task is defined as efficiently locating all targets within the shortest possible timeframe. As depicted in Fig. \ref{figure: multi}, three UAVs are initialized at random positions, and the best region for each UAV is dynamically updated based on the map and the positions of other UAVs. Additionally, each UAV employs the proposed active search strategy to locate the target within its designated region. The simulation results reveal that our method successfully achieves adaptive and well-balanced workload distribution, thereby minimizing the overall search time.

Some other cooperative search methods with multiple UAVs are also implemented for comparison. We also conduct comparisons in three scenes, taking three UAVs as an example. The path length of each UAV and total working time is recorded. As shown in Table. \ref{table: multi}, our method outperforms the other methods in terms of total working time across most scenes. Notably, we combine the proposed Voronoi division method with GM search and HLM search. This combination results in substantial reductions in working time compared to existing methods. Moreover, we observed that the path length of each UAV is similar, indicating an effective workload assignment achieved by our method.

\begin{table*}[htbp!]
\caption{\textbf{Quantitative comparison results on three scenarios}}
\label{table: multi}
\begin{center}
\resizebox{\textwidth}{!}{\begin{tabular}{l c c c c c c}
\toprule
\multirow{2}{*}{Methods} & \multicolumn{2}{c}{Scene 1}  & \multicolumn{2}{c}{Scene 2}  & \multicolumn{2}{c}{Scene 3}\\\cline{2-7}
& Path length of each UAV(m) & working time (s)  & Path length of each UAV(m) & working time (s) & Path length of each UAV(m) & working time (s) \\
\midrule
Zig-zig & 1265, 1265, 1266 & 1266 & 2276, 2276, 2276& 2276 & 357, 357, 358 & 358\\
YAO\cite{Yao2021} & 357, 366, 353 & 366 & 222, 222, 175 & 222 & 165, 195, 175 & 195\\
Manjanna\cite{Manjanna2022} & 249, 249, 249 & 249 & 279, 279, 279& 279 & 221, 221, 221 & 221\\
\hline
Voronoi+GM & 111, 197, 133 & 193 & 150, 150, 120& \textbf{150} & 201, 188, 174 & 201\\

Voronoi+HLM & 156, 156, 156 & 156 & 160, 160, 160& 160 & 173, 173, 166 & 173\\
Ours & 129, 129, 123 & \textbf{129} & 125, 176, 119& 176 & 126, 170, 119 & \textbf{170}\\
\bottomrule
\end{tabular}}
\end{center}
\vspace{-2mm}
\end{table*}
\subsection{Scalability Analysis}
\begin{figure}[!htb]
  \centering
  \includegraphics[width=\hsize]{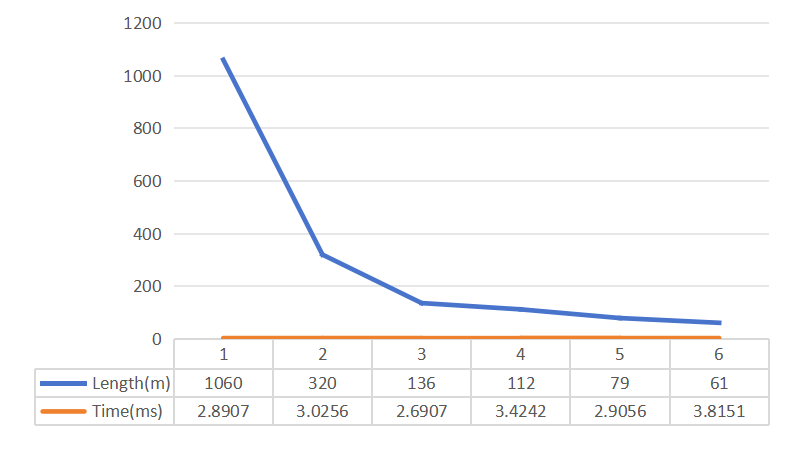}
  \caption{Scalability analysis about the different number of UAVs. The blue line represents the maximum path length in UAVs. The yellow line represents the computation time.}
  \label{figure: scalability}

\end{figure}

To evaluate the scalability of our method, we conducted an analysis using varying numbers of UAVs. We record the maximum path length and computation time for each step. As depicted in Figure \ref{figure: scalability}, it is obvious that as the number of UAVs increases, the maximum search path significantly decreases. Furthermore, due to the fully distributed property of our method, the computation time remains unaffected by the number of UAVs. This scalability analysis highlights the applicability of our method to large-scale collaborative search tasks.

%%%%%%%%%%%%%%%%%%%%%%%%%%%%%%%%%%%%%%%%%%%%%%%%%%%%%%%
\section{Conclusion}
\label{sec: Conclusion}
This paper introduces a cooperative search algorithm that involves multiple UAVs. The proposed method adopts a hierarchical framework, leveraging Voronoi-based region division to ensure balanced workloads for each UAV. Additionally, each UAV incorporates both global and local information to facilitate efficient search within its designated region. To validate the effectiveness of our approach, we conduct several comparisons between active search methods with a single UAV and cooperative search methods with multiple UAVs. These comparisons demonstrate that our method exhibits high efficiency and scalability and achieves good workload assignment. Finally, the proposed algorithm can be extended to large-scale collaborative tasks, which is of great significance to the application of environment monitoring and survey, as well as search and rescue activities.


\begin{thebibliography}{99}

\bibitem{Bouras2022}
A. Bouras, Y. Bouzid, and M. Guiatni, Multi-UAV Coverage Path Planning for Gas Distribution Map Applications, in \emph{Unmanned Systems} 2022, Vol. 10, No. 03, pp. 289-306.

\bibitem{Martinez2022}
J. Martinez-Carranza and L. O. Rojas-Perez, Warehouse Inspection with an Autonomous Micro Air Vehicle, in \emph{Unmanned Systems} 2022, Vol. 10, No. 04, pp. 329-342.

\bibitem{Flaspohler2019}
G. Flaspohler, V. Preston, A. P. M. Michel, Y. Girdhar, and N. Roy, Information-Guided Robotic Maximum Seek-and-Sample in Partially Observable Continuous Environments, in \emph{IEEE Robotics and Automation Letters} 2019, Vol. 4, pp. 3782-3789.

\bibitem{Nawaz2023}
F. Nawaz and M. Ornik, Multi-agent Multi-target Path Planning in Markov Decision Processes, in \emph{IEEE Transactions on Automatic Control} 2023, pp. 1-15.

\bibitem{Igoe2022}
C. Igoe, R. Ghods, and J. Schneider, Multi-Agent Active Search: A Reinforcement Learning Approach, in \emph{IEEE Robotics and Automation Letters} 2022, vol. 7, pp. 754-761.

% \bibitem{Banerjee2023}
% A. Banerjee, R. Ghods and J. Schneider, Multi-Agent Active Search using Detection and Location Uncertainty, in \emph{Proceedings of 2023 IEEE International Conference on Robotics and Automation} London, United Kingdom, 2023, pp. 7720-7727.

\bibitem{Manjanna2017}
S. Manjanna and G. Dudek, Data-driven selective sampling for marine vehicles using multi-scale paths, in \emph{Proceedings of 2017 IEEE/RSJ International Conference on Intelligent Robots and Systems (IROS)}, Vancouver, BC, Canada, 2017, pp. 6111-6117.
	.
\bibitem{Kapoutsis2017}
A. C. Kapoutsis, S. A. Chatzichristofis and E. B. Kosmatopoulos, DARP: Divide Areas Algorithm for Optimal Multi-Robot Coverage Path Planning, in \emph{Journal of Intelligent and Robotic Systems} 2017, vol. 86, pp. 663-680.

\bibitem{Schwager2017}
M. Schwager, P. Dames, D. Rus and V. Kumar, A multi-robot control policy for information gathering in the presence of unknown hazards, in \emph{Robotics Research: The 15th International Symposium ISRR, Springer} 2017, vol. 100, pp. 455-472.

\bibitem{Khan2015}
A. Khan, E. Yanmaz and B. Rinner, Information exchange and decision making in micro aerial vehicle networks for cooperative search, in \emph{IEEE Transactions on Control of Network Systemsr} 2015, vol. 2, pp. 355-347.

\bibitem{Hollinger2014}
G. A. Hollinger and G. S. Sukhatme, Sampling-based robotic information gathering algorithms, in \emph{The International Journal of Robotics Researchy} 2014, vol. 33, pp. 1271–1287. 

\bibitem{Kuhlman2017}
M. J. Kuhlman, M. W. Otte, D. Sofge and S. K. Gupta, Maximizing mutual information for multipass target search in changing environments, in \emph{Proceedings of 2017 IEEE International Conference on Robotics and Automation}, 2017, Singapore, pp. 4383-4390.

\bibitem{Cao2021}
C. Cao, H. Zhu, H. Choset and J. Zhang, TARE: A Hierarchical Framework for Efficiently Exploring Complex 3D Environments, in \emph{Proceedings of Robotics: Science and Systems} 2021, vol. 5.

\bibitem{Dharmadhikari2020}
M. Dharmadhikari, T. Dang, L. Solanka, J. Loje, H. Nguyen, N. Khedekar, et al. Motion Primitives-based Path Planning for Fast and Agile Exploration using Aerial Robots, in \emph{2020 IEEE International Conference on Robotics and Automation (ICRA)} 2020, Paris, France, pp. 179-185.

\bibitem{Dharmadhikari2023}
M. Dharmadhikari and K. Alexis, Semantics-aware Exploration and Inspection Path Planning, in \emph{2023 IEEE International Conference on Robotics and Automation (ICRA)} 2023, London, UK.

\bibitem{Cieslewski2017}
T. Cieslewski, E. Kaufmann and D. Scaramuzza, Rapid exploration with multi-rotors: A frontier selection method for high speed flight, in \emph{Proceedings of 2017 IEEE/RSJ International Conference on Intelligent Robots and Systems (IROS)}, 2017, Vancouver, Canada, pp. 2135-2142.

\bibitem{Lu2020}
L. Lu, C. Redondo and P. Campoy, Optimal Frontier-Based Autonomous Exploration in Unconstructed Environment Using RGB-D Sensor, in \emph{Sensors}, 2020, vol. 20, no. 22, pp 6507.

\bibitem{Bircher2016}
A. Bircher, M. Kamel, K. Alexis, H. Oleynikova and R. Siegwart, Receding Horizon "Next-Best-View" Planner for 3D Exploration, in \emph{Proceedings of 2016 IEEE International Conference on Robotics and Automation (ICRA)}, 2016, Stockholm, Sweden, pp. 1462-1468,

\bibitem{Gelenbe1997}
E. Gelenbe, N. Schmajuk, J. Staddon and J. Reif, Autonomous search by robots and animals: A survey, in \emph{Robotics and Autonomous Systems}, 1997, vol. 22, no. 1, pp. 23-34.

\bibitem{Meghjani2016}
M. Meghjani, S. Manjanna and G. Dudek, Multi-target rendezvous search, in \emph{Proceedings of 2016 IEEE/RSJ International Conference on Intelligent Robots and Systems (IROS)}, 2016, Daejeon, Korea, pp. 2596-2603.

\bibitem{Rückin2022}
J. Rückin, L. Jin and M. Popović, Adaptive Informative Path Planning Using Deep Reinforcement Learning for UAV-based Active Sensing, in \emph{Proceedings of 2022 International Conference on Robotics and Automation (ICRA)}, 2022, Philadelphia, PA, USA, pp. 4473-4479.

\bibitem{Saulnier2020}
K. Saulnier, N. Atanasov, G. J. Pappas and V. Kumar, Information Theoretic Active Exploration in Signed Distance Fields, in \emph{Proceedings of 2020 IEEE International Conference on Robotics and Automation (ICRA)}, 2020, Paris, France, pp. 4080-4085.

\bibitem{Manjanna2022}
S. Manjanna, M. A. Hsieh and G. Dudek, Scalable multirobot planning for informed spatial sampling, in \emph{Autonomous Robots}, 2022, vol. 46, no. 7, pp. 817-829

\bibitem{Yao2021}
P. Yao, L. Qiu, J. Qi and R. Yang, AUV path planning for coverage search of static target in ocean environment, in \emph{Ocean Engineering}, 2021, vol. 241, pp. 110050.

\bibitem{Ebert2022}
J. T. Ebert, F. Berlinger, B. Haghighat and R. Nagpal, A Hybrid PSO Algorithm for Multi-robot Target Search and Decision Awareness, in \emph{Proceedings of 2022 IEEE/RSJ International Conference on Intelligent Robots and Systems (IROS)}, 2022, Kyoto, Japan, pp. 11520-11527.

\bibitem{Ghods2021}
R. Ghods, W. J. Durkin and J. Schneider, Multi-agent active search using realistic depth-aware noise model, in \emph{Proceedings of 2021 IEEE International Conference on Robotics and Automation (ICRA)}, 2021, Xi'an, China, pp. 9101-9108.

\bibitem{Gao2023}
 C. Gao, W. Ding, Z. Zhao and B. M. Chen, Energy-optimal trajectory-based traveling salesman problem for multi-rotors unmanned aerial vehicle, in \emph{Proceedings of the 62nd IEEE Conference on Decision and Control}, 2023, Singapore, pp. 6110–6115.
 
\bibitem{Koyama2020}
A. Koyama, T. Hirata, Y. Kawahara, H. Iyooka, H. Kubozono, N. Onikura, S. Itaya, T. Minagawa, Habitat suitability maps for juvenile tri-spine horseshoe crabs in Japanese intertidal zones: A model approach using unmanned aerial vehicles and the Structure from Motion technique, in \emph{PLOS ONE}, 2020.



\end{thebibliography}
\end{document}